%% file: swin-tuna.tex
\documentclass{article}

\usepackage{PRIMEarxiv}

\usepackage[utf8]{inputenc} 
\usepackage[T1]{fontenc}    
\usepackage{pdfpages}
\usepackage{amssymb}
\usepackage{amsmath}
\usepackage{bm}
\usepackage{lipsum}
\usepackage{multirow}
\usepackage{float}
\usepackage{subfigure}
\usepackage{bbding}
\usepackage{geometry}
\usepackage{graphicx}
\usepackage{makecell}

\usepackage{macros}

\author{
  Haotian Chen \\
  Jiangnan University \\
  Wuxi\\
  \texttt{6233111019@stu.jiangnan.edu.cn} \\
   \And
  Zhiyong Xiao* \\
  Jiangnan University \\
  Wuxi\\
  \texttt{zhiyong.xiao@jiangnan.edu.cn} \\
}

\title{\mytitle}

\begin{document}
\maketitle

\input{sections/1_abstract}

\input{sections/2_introduction}
\input{sections/3_related_work.tex}
\input{sections/4_method}
\input{sections/5_experiments.tex}

\input{sections/6_conclusion.tex}

\bibliographystyle{unsrt}
\bibliography{ref}
\end{document}

%% file: sections/1_abstract.tex
\begin{abstract}
%
%
%
%
%
%
In the field of food image processing, efficient semantic segmentation techniques are crucial for industrial applications. However, existing large-scale Transformer-based models (such as FoodSAM) face challenges in meeting practical deployment requirements due to their massive parameter counts and high computational resource demands. This paper introduces TUNable Adapter module (Swin-TUNA), a Parameter Efficient Fine-Tuning (PEFT) method that integrates multiscale trainable adapters into the Swin Transformer architecture, achieving high-performance food image segmentation by updating only 4\% of the parameters. The core innovation of \mynet lies in its hierarchical feature adaptation mechanism: it designs separable convolutions in depth and dimensional mappings of varying scales to address the differences in features between shallow and deep networks, combined with a dynamic balancing strategy for tasks-agnostic and task-specific features. Experiments demonstrate that this method achieves mIoU of 50.56\% and 74.94\% on the FoodSeg103 and UECFoodPix Complete datasets, respectively, surpassing the fully parameterized FoodSAM model while reducing the parameter count by 98.7\% (to only 8.13M). Furthermore, \mynet exhibits faster convergence and stronger generalization capabilities in low-data scenarios, providing an efficient solution for assembling lightweight food image.

\noindent The codes are available at \github.
\end{abstract}

%% file: sections/2_introduction.tex
\section{Introduction}
In today's era of rapid development of digitalization and intelligence, food image processing technology is gradually becoming a hotspot and focus of research in food science, computer vision, and related fields. In the food field, image classification, target detection, and semantic segmentation are the most widely used methodologies\cite{SONG2025104982}. Fine-grained segmentation of food images plays a crucial role as one of the key technologies \cite{FGFoodNet}.

Food image segmentation is a fundamental and challenging task in computer vision, which aims to accurately distinguish food pixels from background pixels in an image. Fine-grained segmentation, on the other hand, goes further, not only to distinguish food from the background, but also to make fine divisions within the food to recognize the different components of the food, such as identifying peel, pulp and core in fruits, or segregating lean meat, adipose tissue, and fascia in meat products. This fine-grained segmentation has irreplaceable and important value for many practical applications \cite{qyh,Wang2025}. Hybrid Parallel Genetic Algorithm is particularly suitable for automated segmentation tasks involving high-dimensional, high-resolution, and multi-modal images \cite{HPGA2021}. 

\input{figures/params_mIoU}
Transformer-based models \cite{XIAO2022, XIAO2024} demonstrate the great potential of the attention architecture for processing food images. Recent models such as FoodSAM\cite{foodsam}, achieve excellent results by combining the image segmentation base model, the SAM (Segment Anything Model) \cite{kirillov2023segany}, but in practice, achieving a balance between speed, segmentation, and recognition accuracy remains challenging for FoodSAM. FoodSAM is based on a SAM 2b model, and an additional coarse mask generation model based on the target task requires training. Training a model to perform a specific food-related task requires significant computing power and time. Meanwhile, techniques applied to food (e.g., video segmentation, food composition analysis, and assembly line quality inspection) often require cheap, high-volume computing resources. Although models such as FoodSAM can achieve high accuracy, they are not fast enough to meet industrial requirements.

PEFT methods have demonstrated promising results in achieving rapid performance convergence with minimal parameter updates. These approaches enable adaptation of pre-trained models through selective tuning of limited parameters, often matching or exceeding full fine-tuning performance. While PEFT has gained widespread adoption in NLP \cite{nlp1, nlp2, nlp3, nlp4, nlp5}, its application in computer vision - particularly for food image segmentation - remains underexplored. Existing attempts, such as \cite{sahay2024mopeftmixtureofpeftssegmentmodel}, have yet to surpass full fine-tuning baselines in segmentation accuracy.

To address these challenges, this paper present \mynet, an innovative plug-and-play PEFT module that integrates multi-scale trainable modules into Swin-Transformer\cite{swin} architectures. Our approach maintains model efficiency by freezing the majority of pre-trained parameters while only updating the segmentation head and injected adapters. Experimental results demonstrate that \mynet achieves \foodsegmiou ~mIoU on FoodSeg103\cite{wu2021foodseg} and 74.94\% mIoU on UECFoodPix Complete \cite{uecfoodpixcomplete}, outperforming existing models while training merely \parameters ~of total parameters. 

The main contributions of this paper are as follows:

\begin{enumerate}
\item First, it demonstrates that efficient fine-tuning methods can achieve or even surpass full fine-tuning on small sample datasets, such as food image segmentation, with a small number of parameters.

\item It proposes Swin-TUNA, a lightweight, plug-and-play module that uses adaptive layer-related parameters to efficiently retain knowledge from pre-training weights for downstream tasks.

\item Third, extensive experiments were designed to demonstrate that Swin-TUNA outperforms other PEFT models on FoodSeg103 and UECFoodPix Complete. Swin-TUNA even outperforms the full-volume fine-tuning model, FoodSAM, while adjusting only 4\% of the parameters. To the best of our knowledge, this is the first time the PEFT method has been used for food image segmentation while achieving state-of-the-art (SOTA) results.
\end{enumerate}

%% file: figures/params_mIoU.tex
\begin{figure}[ht]
    \centering
    \includegraphics[width=8cm]{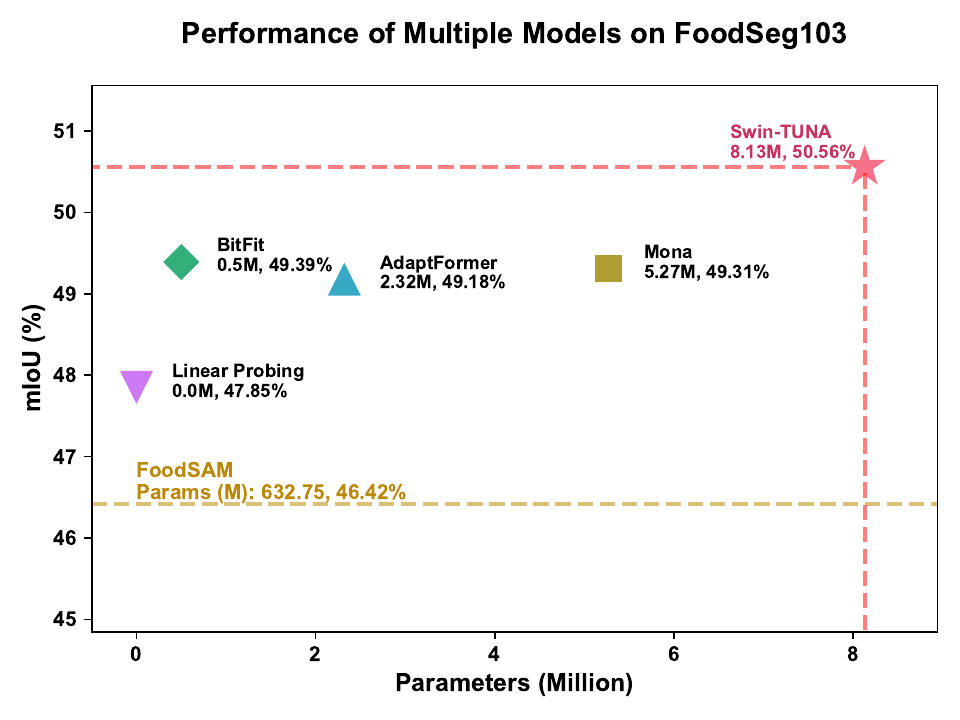}
    \caption{Comparison results of Swin-TUNA, FoodSAM, and other PEFT methods in terms of parameter count and mIoU. The yellow dashed line represents the overall fine-tuning of the FoodSAM model, and \mynet surpassed the FoodSAM (632.75M) only with 8.13M trainable parameters. The experimental results demonstrate that efficient fine-tuning methods can also outperform traditional full fine-tuning.}
\end{figure}

%% file: sections/3_related_work.tex
\section{Related Work}
\smalltitle{Previous Work on Food Image Processing}  
Since Vision Transformer (ViT)\cite{vit} breakthroughly introduced the pure Transformer architecture into the field of computer vision, vision models based on the self-attention mechanism have demonstrated performance advantages over traditional CNNs \cite{Xiao2019,Qian2020,CMPB2021,BSPC2021,Liu2023,BEL2024}. Many models constructed by stacking multi-layer Transformer blocks (e.g., Swin Transformer\cite{swin}, PVT\cite{pvt}) have achieved better results in many fields such as image classification, target detection, semantic segmentation, image super-resolution, video tracking, etc., and have constantly set new records in benchmark tasks such as ImageNet classification and COCO target detection. Meanwhile, a large number of large-scale pre-training weights enable these models to perform well on downstream tasks by means of migration learning. By compressing hundreds of spectral bands into just a handful of feature maps through hyperspectral dimensionality reduction and feeding them into classical or deep-learning models, we can achieve fast and accurate pixel-level segmentation for food quality inspection and nutritional assessment \cite{RSL2014, Xiao2014}.

In the context of food image classification, several noteworthy methods have emerged. 
\cite{gxl} proposed an improved method of Vision Transformer combining data enhancement and feature enhancement (ALSMViT). The method effectively solves the classification difficulties of traditional methods when dealing with food products with similar shapes but different nutritional values through the multi-layer perception mechanism of AugmentPlus data enhancement strategy, LayerScale deep architecture and local enhancement of features.
Similarly, \cite{lrk} proposed a FoodCSWin method based on CSWin Transformer with diffusion model, which significantly improves the recognition accuracy of similar nutritionally similar but visually different food products through DiffAugment data augmentation and LFDB-block local feature enhancement module (ChineseFoodNet 85.67\%, VireoFood172 94.11\%), providing a high-precision solution for image recognition-based dietary assessment.
Furthermore, \cite{dg}proposed a fine-grained food image recognition method Swin-DR based on Swin Transformer, which enhances the local feature representation by incorporating Deep Residual Convolution Module (DrConvBlock) and designing the end-to-end generalized classifier MLP-GD, which achieves 81.07\% accuracy on FoodX-251 and UEC Food-256 datasets, achieving 81.07\% and 82.77\% accuracy on FoodX-251 and UEC Food-256, respectively, which outperforms existing self-supervised methods.

The prevailing food image segmentation techniques are predominantly founded on Transformer models. In their seminal work, Wu et al.\cite{wu2021foodseg} constructed a new dataset, FoodSeg103, and proposed ReLeM, a multimodal knowledge migration method. This innovative approach aims to enhance the performance of the model by fusing linguistic and visual features. OVFoodSeg\cite{wu2024ovfoodsegelevatingopenvocabularyfood} employs a comparable concept, utilizing a two-stage training approach. Initially, images are pre-trained on a text learner (FoodLearner), aligning visual and textual features. Subsequently, the segmentation task is adapted, yielding favorable outcomes on FoodSeg103. CANet\cite{canet} has been developed to optimize feature extraction efficiency and global dependency capture by designing a Cross-Spatial Attention (CSA) module.

What sets FoodSAM\cite{foodsam} apart from other models is its integration of the foundational vision model Segment Anything Model (SAM)\cite{kirillov2023segany} with food images for the first time. The methodology of the study began with the training of a semantic segmentation model capable of generating preliminary masks to obtain label information. Subsequently, the SAM's unsupervised mask generation technique was employed to process the masks, thereby achieving SOTA performance in three domains: semantic segmentation, instance segmentation, and target detection for food images.

In addition to Transformer-based approaches, hybrid models that combine CNNs and Transformers have also demonstrated significant potential. For instance, \cite{ly} proposed a dual-branch structure network (FDSNet) that integrates Swin Transformer and CNNs. The shallow branch processes full-size, high-frequency residual images to extract spatial details, while the deep branch processes downsampled images to capture semantic information. The study also introduced an advanced multiscale feature fusion module to enhance performance, thereby demonstrating the potential of hybrid architectures for addressing food image segmentation tasks.

Notably, the number of parameters in typical visual Transformer models shows exponential growth. For example, comparing the CNN method in FoodSeg103 (25.6 million parameters) and FoodSAM (632.75 million parameters), the mean intersection over union (mIoU) improves by 26.14\%, from 36.8\% to 46.42\%, but the number of parameters increases by a factor of 24. This scale expansion significantly increases computational demand despite improving characterization capability. For instance, SAM required 256 A100 graphics cards for 56 hours of training\cite{kirillov2023segany}, far surpassing the computational capacity of most research organizations.

Additionally, when full fine-tuning is performed on a large-scale pre-trained visual Transformer for migration learning, the model is susceptible to catastrophic forgetting\cite{goodfellow2015empiricalinvestigationcatastrophicforgetting, mccloskey1989catastrophic}. This means that it loses important knowledge gained in the pre-training phase when training a new task. This phenomenon is particularly significant in scenarios with small sample datasets (e.g., food datasets and medical images) \cite{lin2024mitigatingalignmenttaxrlhf, dutt2024parameterefficientfinetuningmedicalimage}. For instance, on the HAM10000 dataset, full-volume fine-tuning yields a substantial decline in F1 scores when the data set is reduced to less than 50\%. This seriously constrains the model's ability to continuously learn and its practical application value.

\input{figures/vis_seg}
\smalltitle{Advancements and Limitations of PEFT}
To cope with the above challenges, the PEFT approach has emerged. This technical route first made a breakthrough in the field of NLP, with representative works such as Adapter\cite{adapter}, which achieves full-parameter fine-tuning effect by inserting a two-layer MLP structure between each Transformer layer at a cost of only 3.6\% increase in the number of parameters on the GLUE benchmark. The subsequent AdapterFusion\cite{adapterfusion} proposed a structure for sharing multi-task information that combines knowledge from multiple tasks to achieve parameter sharing among multiple tasks, effectively mitigating the problems of catastrophic forgetting, inter-task interference, and training instability. These works lay the foundational paradigm of PEFT for NLP tasks and demonstrate significant advantages in terms of parameter efficiency (typically only 0.1\%-5\% trainable parameters are required) and knowledge retention.

In PEFT, the Adapter method belongs to a category that introduces extra parameters. Although the Adapter method is very efficient, the additional structure introduces extra parameters, which affects the speed of reasoning to some extent. In contrast, there are methods that do not require the introduction of additional parameters in reasoning. One example is linear probing, a fairly simple method that trains different tasks by freezing all the parameters in the backbone and training only the segmentation and classification headers. A slightly more complex method is LoRA\cite{lora}, which fine-tunes the model by adding trainable low-rank matrices next to the pre-trained weights through low-rank decomposition. These matrices are then merged into the model, thus avoiding the introduction of additional parameters. BitFit\cite{bitfit} trains only the bias parameters in the model, which account for about 0.1\% of the total parameters. Prompt-based methods \cite{nlp1, nlp3, nlp4} directly modify the model's prompts to fine-tune without altering the network itself, although their effectiveness is limited.

These methods were soon introduced to the vision domain. VPT\cite{vpt}proposes a method to achieve 95\% full fine-tuning performance in the ADE20K semantic segmentation task with 2\% trainable parameters by modifying the inputs to the model and generating the appropriate Prompt vectors for fine-tuning for each input image. \cite{touvron2022thingsknowvisiontransformers} performs migration learning by only fine-tuning the Attention module. AdaptFormer\cite{adaptformer} draws on the Adapter method in NLP by modifying the MLP in the Transformer Block to introduce trainable parameters to fine-tune the model. Similar to AdaptFormer is Mona\cite{mona}, which breaks new ground in several areas such as semantic segmentation, target detection, and image classification by introducing multi-cognitive visual filters.

However, the performance of existing PEFT methods for food image segmentation is still unsatisfactory. Methods that do not require additional trainable modules, such as LoRA, BitFit, and Attention Tuning, have high training efficiency. Nevertheless, the accuracy cannot surpass that of full fine-tuning. Prompt-based methods are advantageous when the number of tokens is small, but accuracy decreases for tasks with a large number of tokens\cite{adaptformer}. Additionally, most existing adapter methods use a homogenized parameter insertion strategy (e.g., adding the same structure adapter in each transformer block), ignoring the hierarchical difference between deep semantic features and shallow texture features. These issues are especially relevant in food image segmentation tasks because the fine-grained textures and complex spatial relationships of ingredients necessitate more precise fine-tuning strategies. These unresolved issues provide a theoretical basis and room for innovation in researching this paper's method.

%% file: figures/vis_seg.tex
\begin{figure*}[ht]
    \begin{minipage}{\linewidth}
        \centering
        \includegraphics[width=2.8cm]{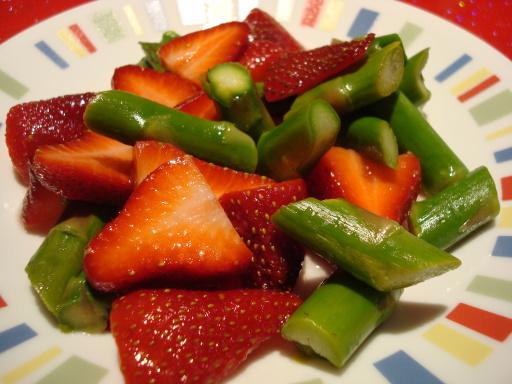}
        \includegraphics[width=2.8cm]{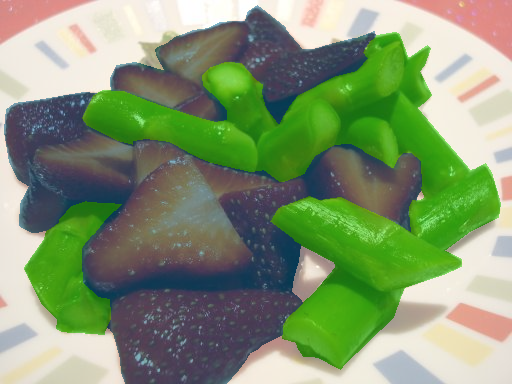}
        \includegraphics[width=2.8cm]{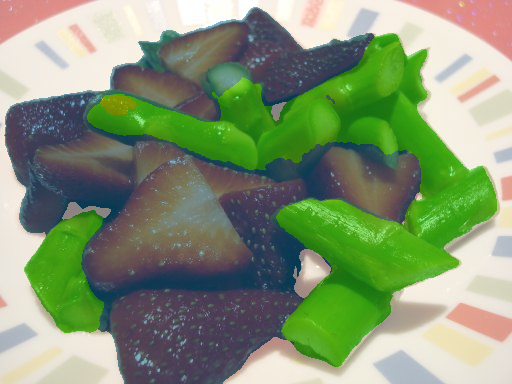}
        \includegraphics[width=2.8cm]{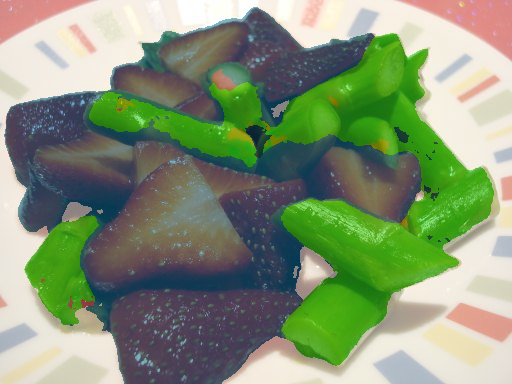}
        \includegraphics[width=2.8cm]{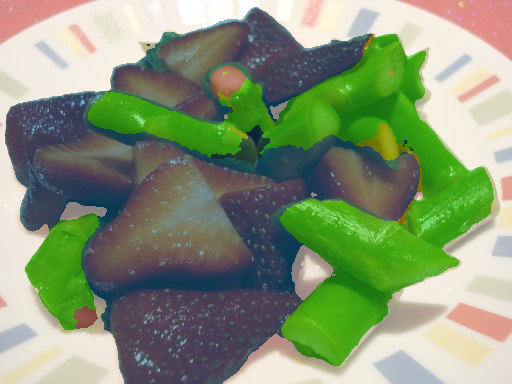}
    \end{minipage}
    
    \begin{minipage}{\linewidth}
        \centering
        \begin{minipage}{2.8cm}
            \centering
            \includegraphics[width=2.8cm]{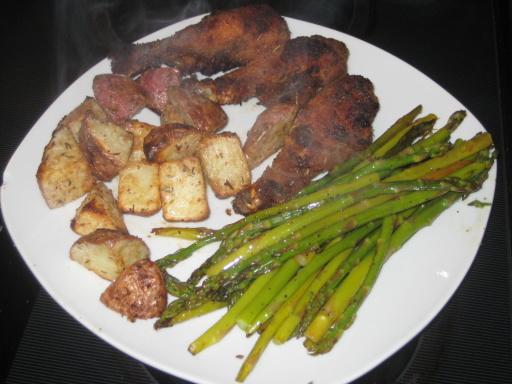}
            \\
            \footnotesize
            Original Image
        \end{minipage}
        \begin{minipage}{2.8cm}
            \centering
            \includegraphics[width=2.8cm]{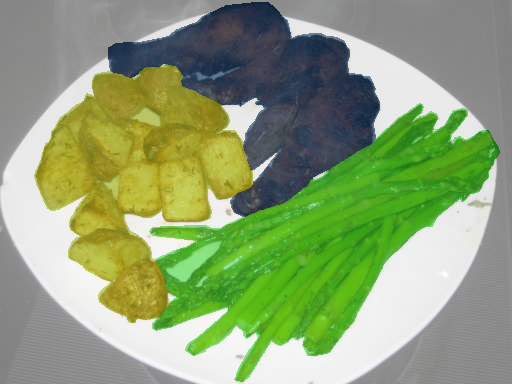}
            \\
            \footnotesize
            Ground Truth
        \end{minipage}
        \begin{minipage}{2.8cm}
            \centering
            \includegraphics[width=2.8cm]{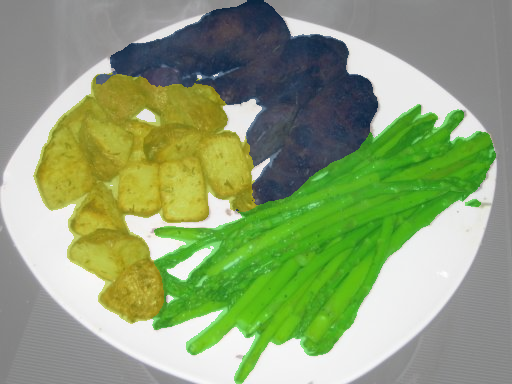}
            \\
            \footnotesize
            \mynet
        \end{minipage}
        \begin{minipage}{2.8cm}
            \centering
            \includegraphics[width=2.8cm]{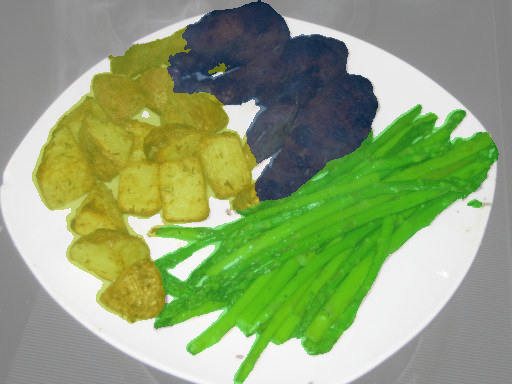}
            \\
            \footnotesize
            AdaptFormer
        \end{minipage}
        \begin{minipage}{2.8cm}
            \centering
            \includegraphics[width=2.8cm]{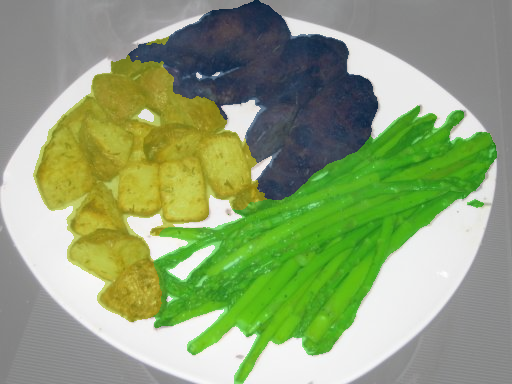}
            \\
            \footnotesize
            BitFit
        \end{minipage}
    \end{minipage}
    \caption{The results of \mynet and other PEFT models in segmenting images on FoodSeg103}
\end{figure*}

%% file: sections/4_method.tex
\section{Method}
This chapter introduces TUNA, a PEFT method that effectively leverages the knowledge of large pre-trained backbones and transfers it to various downstream tasks, requiring only a minimal number of parameters to achieve significant results. The TUNA method applied to the Swin Transformer is referred to as Swin-TUNA.

\input{figures/tuna}

\subsection{Adapter}

Some capable teams and companies release their pre-trained weights on the backbone for other teams to apply to downstream tasks. This process can be represented by the following formula:

\begin{equation}
    w = \arg\min_{w} \mathbb{E}_{(x,y) \sim D_{large}} \left[ \mathcal{L}(f(x; w),y) \right]
    \label{eq.pre_ft}
\end{equation}

\begin{equation}
    w' = \arg\min_{w'} \mathbb{E}_{(x,y) \sim D_{task}} \left[ \mathcal{L}(f(x; w'),y) \right]
    \label{eq.full_ft}
\end{equation}

Here, $D_{large}$ is the pre-training dataset, $D_{task}$ is the dataset for the downstream task, $\mathcal{L}$ is the loss function, $f$ is the model, and $w$ represents the model parameters.

Formula \ref{eq.pre_ft} represents the process of training the backbone to obtain pre-trained weights, while formula \ref{eq.full_ft} indicates the fine-tuning for the downstream task. Even when $D_{task}$ is small, the formula \ref{eq.full_ft} requires adjusting all parameters to optimize the model, often resulting in low training accuracy \cite{dutt2024parameterefficientfinetuningmedicalimage} and issues such as catastrophic forgetting \cite{goodfellow2015empiricalinvestigationcatastrophicforgetting}\cite{mccloskey1989catastrophic}.

In the domain of PEFT, the Adapter method is a straightforward yet effective approach. By introducing and adjusting only additional parameters, the model can achieve results that meet or exceed those of the full parameter adjustment in formula \ref{eq.full_ft}. The training process of the Adapter method is as follows:

\begin{equation}
    \theta^* = \arg\min_{\theta} \mathbb{E}_{(x, y) \sim D_{task}}\left[ \mathcal{L}(f(x; w, \theta)) \right]
    \label{eq.delta}
\end{equation}

Here, $\theta^*$ represents the additional parameters that are introduced and trained. Typically, $\theta^*$ is significantly smaller than $w'$.

\subsection{\mynet}

\smalltitle{Architecture} \mynet is a variant of the Adapter method in PEFT. The Adapter method typically consists of a downsampling module, a nonlinear function, and an upsampling module. The design philosophy of the Adapter paradigm is to minimize the number of introduced parameters while closely approximating or surpassing full fine-tuning. To ensure \mynet enhances training efficiency with minimal parameters, it employs convolutions as the nonlinear function. Unlike standard convolutions, \mynet utilizes a combination of depth-wise convolutions and 1x1 convolutions to replace a single convolutional kernel, achieving comparable results with fewer parameters.

The specific architecture of \mynet is illustrated in the figure. \mynet is injected into the Swin Transformer Block, modifying the incoming parameters and computing them back into the computation chain. Specifically, the input to \mynet first undergoes a downsampling module, followed by an $n \times n$ depth-wise convolution, and is then mapped to the output distribution using a $1 \times 1$ convolution. According to ConvNeXt, the mapping before upsampling aids in computational stability. Subsequently, an upsampling module maps the input channels back to the input channels and is followed by the GeLU activation function.

Empirical studies have shown that allowing Dropout in the backbone to function helps prevent the model from overfitting. Therefore, during training, all Dropout layers must also operate normally in addition to \mynet participating in the training process.

The entire computational process of the \mynet module can be expressed as follows:

\begin{equation}
\begin{aligned}
    &X_{down} = F_{down}(X_{in})\\
    &X_{up} = F_{up}(Conv(X_{down}) + X_{down})\\
    &X_{out} = Dropout(\sigma(X_{up})) + X_{in}\\
    \label{eq.tuna}
\end{aligned}
\end{equation}

Here, $X_{in}, X_{out}$ represent the inputs and outputs of the module, while $X_{down}, X_{up}$ correspond to the downsampling and upsampling modules, respectively. $\sigma$ is the activation function, and Conv represents the depth-wise convolution.

The computational flow of a Swin Transformer Block is ultimately represented as follows. For descriptive purposes, the \hot{red parts} of the formula represent the additional parameters introduced to the Transformer layer. The black parts represent the processes inherent to the original Transformer layer. Therefore, only the red parts participate in training and update their parameters according to the gradient.

\begin{equation}
\begin{aligned}
    &{{\hat{\bf{z}}}^{l}} = \text{W-MSA}\left( {\text{LN}\left( {{{\bf{z}}^{l - 1}}} \right)} \right) + {\bf{z}}^{l - 1},\\
    &{{\bf{z}}^l} = 
    \hot{{\bf{s}}_1^l} \otimes \left(\text{MLP}\left( {\text{LN}\left( {{{\hat{\bf{z}}}^{l}}} \right)} \right) 
    \oplus {{\hat{\bf{z}}}^{l}}\right)\\
    &~~~+ \hot{{\bf{s}}_2^{l}} \times \hot{\text{TUNA}^{l}}({\bf{z}}^{l-1}),\\
    &{{\hat{\bf{z}}}^{l+1}} = \text{SW-MSA}\left( {\text{LN}\left( {{{\bf{z}}^{l}}} \right)} \right) + {\bf{z}}^{l}, \\
    &{{\bf{z}}^{l+1}} = 
    \hot{{\bf{s}}_1^{l+1}} \otimes \left(\text{MLP}\left( {\text{LN}\left( {{{\hat{\bf{z}}}^{l+1}}} \right)} \right) \oplus {{\hat{\bf{z}}}^{l+1}}\right)\\
    &~~~~~~+ \hot{{\bf{s}}^{l+1}_2} \times \hot{\text{TUNA}^{l+1}}({\bf{z}}^{l-1})\\
    \label{eq.swin_tuna}
\end{aligned}
\end{equation}

where $\otimes$ denotes element-wise multiplication and $oplus$ residual addition. To balance the task-agnostic features (generated by the original frozen branch) and the task-specific features, \mynet introduces two trainable parameters, $s_1$ and $s_2$, where $s_1$ is initialized to 1e-6 and $s_2$ to 0.

\smalltitle{Hierarchical Feature Adaptation} Previous studies have shown that shallow network features in deep networks can capture high-frequency information locally \cite{simonyan2015deepconvolutionalnetworkslargescale, zeiler2013visualizingunderstandingconvolutionalnetworks}, while deep network features better represent global information, possessing larger receptive fields and rich semantic information \cite{he2015deepresiduallearningimage}. To enable the model to better integrate high-frequency and low-frequency features for improved semantic information capture, this paper proposes a straightforward and efficient method: designing distinct parameters for different layers of the Swin Transformer.

Specifically, the Swin Transformer is a four-layer model, indexed as $[0, 1, 2, 3]$. Higher indices denote deeper networks. To enhance the shallow networks' focus on global information while allowing deep networks to compute local details, the convolution kernels in \mynet are designed to be smaller in the shallow layers and larger in the deeper layers. In practice, the convolutional kernel sizes are set to $[7, 5, 5, 3]$.

Unlike standard convolutions, the convolution operations used in \mynet are depth-wise convolutions. This means each filter in the convolution operates independently on the corresponding channel's features. This reduces the number of parameters and computational load, but it lacks interaction between different channels. Therefore, \mynet employs point-wise convolutions after depth-wise convolutions to upsample the features.

Additionally, the dimensions of the mappings for downsampling and upsampling also vary across layers. In practice, the dimension sizes are set to $[64, 64, 96, 192]$.

\smalltitle{Training and Inference} Throughout the training process, only $s_1$, $s_2$, and the TUNA module participate in backpropagation in Formula \ref{eq.swin_tuna}. This means that only the parts marked in red, which are additionally injected into the Swin Transformer layers, and Dropout will participate in training, while the parameters in blue remain frozen. Consequently, only the injected portions of the model will acquire task-specific knowledge, while the backbone retains the substantial knowledge gained from pre-training. For different tasks, the trained models share the backbone parameters.

%% file: figures/tuna.tex
\begin{figure}[ht]
    \centering
    \includegraphics[width=0.5\textwidth]{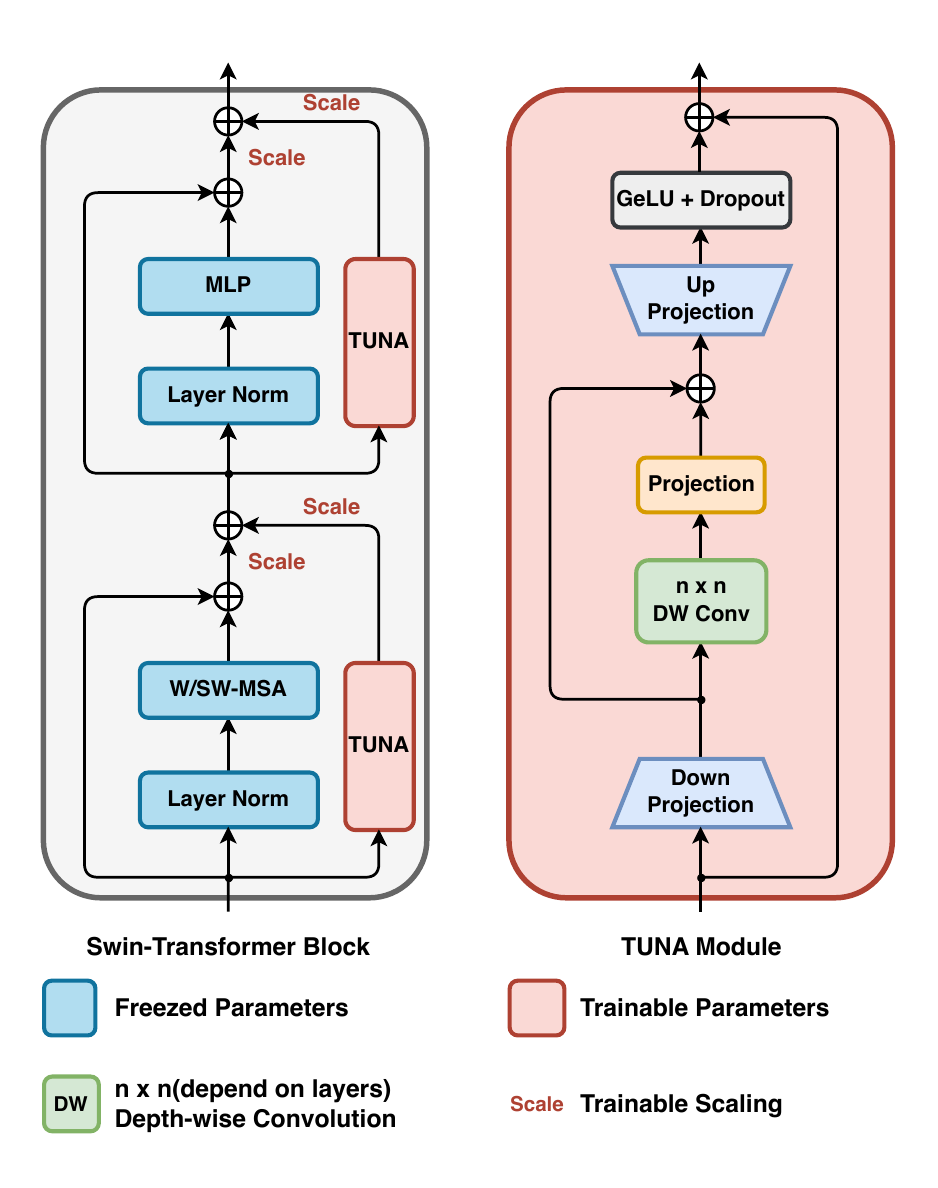}
    \caption{\textbf{Architectural Overview of TUNA.} Left: The proposed \mynet. Right: Detail of TUNA. TUNA follows the paradigm of Adapter\cite{adapter} and consists of an upsampling module, a downsampling module, and a non-linear module (TUNA uses Depthwise convolution). Unlike Adapter, all three modules of TUNA have hierarchical adaptivity, which enables better processing of features at different levels and endows TUNA with the ability to perform more than full fine-tuning.}
    \label{fig:tuna}
\end{figure}

%% file: sections/5_experiments.tex
\section{Experiments}
\input{figures/comparison_of_peft}
In this chapter, experiments were designed to validate the capability of \mynet on different datasets. Section \ref{sec.exp.setup} introduces the relevant basic setup of the experiments, including the datasets used, baseline, model parameter details and the experimental environment. Section \ref{sec.exp.main} compared and analyzed the results of \mynet and baseline  on two common datasets used for segmentation of food images, FoodSeg103 and UECFoodPix Complete, respectively. Section \ref{sec.exp.abl} designed ablation experiments on the structure of \mynet to demonstrate the correctness of the architectural design.

\subsection{Experimental Settings}
\label{sec.exp.setup}
\smalltitle{Evaluation Metrics}
Following standard semantic segmentation protocols, we employ three evaluation metrics:
\begin{itemize}
\item \textbf{mIoU}: Mean intersection-over-union across classes:
\begin{equation}
\text{mIoU} = \frac{1}{C}\sum_{i=1}^{C} \frac{TP_i}{TP_i + FP_i + FN_i}
\end{equation}

\item \textbf{mAcc}: Mean pixel accuracy per class:  
\begin{equation}  
    \text{mAcc} = \frac{1}{C}\sum_{i=1}^{C} \frac{TP_i}{TP_i + FN_i}  
\end{equation}  
  
\item \textbf{aAcc}: Overall pixel accuracy:  
\begin{equation}  
    \text{aAcc} = \frac{\sum_{i=1}^{C}TP_i}{\sum_{i=1}^{C}(TP_i + FP_i)}  
\end{equation}  
\end{itemize}
where $C$ denotes total classes, $TP_i$, $FP_i$ and $FN_i$ represent true positives, false positives, and false negatives for class $i$.

\smalltitle{Baseline}
In this paper, we compare Swin-TUNA with two types of methods that are categorized according to the fine-tuning method, namely (1) full fine-tuning methods, which contain mainstream methods such as FoodSAM. (2) PEFT methods, which include BitFit, Linear Probing without additional parameters; and Mona, AdaptFormer with additional parameters.

\smalltitle{Datasets}
This paper selects two popular food image segmentation datasets. To better investigate data distribution and facilitate training and analysis, two analytical approaches are applied to the datasets: the Resolution Range Ratio and the Area Distribution-based Gini Coefficient.

\begin{itemize}
    \item \textbf{Resolution Range Ratio}: Measures the size span of the dataset by calculating the ratio of the maximum area to the minimum area. This metric reflects the extent of extreme resolution differences among images in the dataset. A larger value indicates greater area disparity.
\begin{equation}
    R_{\text{range}} = \frac{A_{\max}}{A_{\min}}
\end{equation}
where $A_{\max}$ is the largest image area in the dataset, $A_{\min}$ is the smallest image area in the dataset, $A_i = w_i \times h_i$, and $w_i$ and $h_i$ are the width and height of the $i$-th image, respectively.

    \item \textbf{Area Distribution-based Gini Coefficient}: Adapted from the measure of income inequality in economics, it quantifies the unevenness of image area distribution. This coefficient captures the overall shape of the area distribution, rather than solely considering extremes. A larger value indicates a more uneven area distribution.
\begin{equation}
    G = \frac{1}{2n^2\bar{A}} \sum_{i=1}^{n} \sum_{j=1}^{n} |A_i - A_j|
\end{equation}
where $n$ is the total number of images, $A_{(i)}$ is the area of the $i$-th image after sorting by area in ascending order, and $i$ is the index position after sorting.
\end{itemize}

\smalltitle{FoodSeg103\cite{wu2021foodseg}}
FoodSeg103 is a large-scale benchmark for food image segmentation. It is based on Recipe1M and contains 7,118 images and annotations of Western food products. Of these, 4,983 images were used for training and 2,135 for testing. The dataset encompasses annotations for 104 food ingredient categories, with an average of six annotations per image.
The three most common sizes in this dataset are $512 \times 384$, $256 \times 256$, and $3264 \times 2448$; the an average size is $770.9 \times 646.8$. The resolution range ratio is 1094.55, and the area Gini coefficient is 0.76. The analysis results indicate significant variations in image resolution within the dataset, with a substantial proportion of images exhibiting low resolution and a subset of images displaying ultra-high resolution. Additionally, the dataset is predicated on food ingredients as the fundamental segmentation unit, rather than dishes, which creates challenges in segmenting the images.

\noindent \textbf{UECFoodPix Complete\cite{uecfoodpixcomplete}}: The dataset is based on UEC-Food-100, with GrabCut assisting in the generation of bounding boxes, and it has been refined by manual correction. The database contains a total of 10,000 images, of which 9,000 are designated for training purposes and 1,000 are allocated for testing. These images encompass 102 distinct food categories, including well-known Japanese culinary items such as ramen, rice, seafood, desserts, vegetables, and more.
The three most common sizes in this dataset are $640\times480$, $800\times600$, and $480\times360$; the average size is $442.4\times349.1$. The resolution range ratio is 78.90, and the area Gini coefficient is 0.40. The results of the analysis indicate that the dataset exhibits a paucity of data down to the ingredient level. However, this dataset contains a more substantial amount of data and demonstrates greater stability in its resolution, which facilitates more efficient training in comparison to FoodSeg103.

\smalltitle{Implementation Details}
All experiments were conducted on an NVIDIA RTX 3090 GPU using:
\begin{itemize}
\item Framework: MMsegmentation \cite{mmseg2020} with PyTorch 1.13.1
\item Backbone: Swin-L pretrained on ImageNet-22K
\item Optimization: AdamW (lr=1e-4, weight decay=0.01)
\item Training: 90k iterations in UECFoodPix Complete, 100k in FoodSeg103, with cosine annealing
\end{itemize}

\subsection{Main Results}
\label{sec.exp.main}
\input{tables/comp_foodseg}
\input{tables/comp_uec}
\smalltitle{SOTA Comparison}
As can be seen in the Table \ref{tb.foodseg103}, \mynet outperforms the other PEFT methods on FoodSeg103 and eventually outperforms all other methods. Specifically, \mynet outperforms BitFit by \hot{1.17\%} and FDSNet by \hot{3.22\%} on FoodSeg103. As shown in the and Table \ref{tb.uec}, mIoU outperforms Mona by \hot{0.24\%} but is lower than the full fine-tuning model, FDSNet, by \freeze{0.95\%} on UECFoodPix Complete.

Notably, linear probing (0\% backbone tuning) outperforms full fine-tuning baselines by 1.43-11.63\% mIoU on FoodSeg103, confirming PEFT's superiority in low-data regimes. This aligns with findings in medical imaging \cite{dutt2024parameterefficientfinetuningmedicalimage}, where limited training samples ($\leq$5k) favor parameter-efficient approaches.

In summary, the \mynet method demonstrates a high degree of efficiency, outperforming all PEFT methods and meeting or exceeding the full amount of fine-tuning with a mere 8\% of the parameters trained.

\subsection{Efficiency Analysis}
Figures \ref{fig:foodseg103} and \ref{fig:uec} illustrate the loss convergence and mIoU testing curves of \mynet and representative PEFT methods. In Figure  \ref{fig:foodseg103}, it is evident that during training on FoodSeg103, \mynet consistently achieves higher mIoU compared to other PEFT methods, surpassing the peak performance of all other models by the 50,000th iteration.  

As shown in the Figure \ref{fig:uec}, all of the aforementioned PEFT methods, including Mona, which leverages multi-cognitive visual filters, converge less stably on the larger, more complex UECFoodPix Complete dataset. Nevertheless, Swin-TUNA's convergence speed is significantly faster than that of the other PEFT methods.

In summary, \mynet demonstrates superior performance in terms of convergence speed and accuracy when compared to other PEFT methods, with a parameter count of 8.13M, which is marginally higher than the counts of the other methods.

\subsection{Ablation Studies}
\label{sec.exp.abl}
This section conducts a series of ablation experiments to identify which factors contribute to the effectiveness of \mynet and to explore the underlying reasons. All ablation studies in this work are performed on FoodSeg103, using mIoU as the evaluation metric.

\input{figures/module_position}
\smalltitle{Adapter Structure}  
This section compares two different backbone structures of \mynet, as illustrated in Figure \ref{fig:structure}. Experimental results on FoodSeg103 indicate that the parallel structure surpasses the sequential structure by 0.3\% in mIoU. The likely reason is that the sequential structure simply stacks modules, processing features one at a time, whereas the parallel structure allows the Swin and TUNA modules to process the same features differently, with the final output balanced by a scaling vector.

\input{tables/ablation_foodseg}
\smalltitle{Hierarchical Feature Adaptation}
It has been observed that some existing PEFT methods \cite{adaptformer, mona} insert identical structures within Transformer-based layers to fine-tune the model. However, as the network depth increases, certain Transformers (like the Swin Transformer) undergo structural changes, and the semantic information of the features also evolves; therefore, employing a uniform architecture is unreasonable. This section designs four distinct architectural parameter schemes:

\begin{itemize}
    \item Fixed dimension of 64 with a fixed 3x3 convolution; 
    \item Dimensions varying with depth as [64, 64, 96, 192] with a fixed 3x3 convolution; 
    \item Fixed dimension of 64 with convolution sizes varying with depth as 7, 5, 5, 3; 
    \item Dimensions varying with depth as [64, 64, 96, 192] and convolution sizes varying with depth as 7, 5, 5, 3.
\end{itemize}
The experimental results are presented in Table \ref{tb.alb}. The results demonstrate that the modules employing the Hierarchical Feature Adaptation method effectively integrate shallow and deep features, yielding superior performance. Consequently, \mynet ultimately adopts the parameter scheme, where dimensions vary with depth as [64, 64, 96, 192] and convolution sizes vary as [7, 5, 5, 3].
%

%% file: figures/comparison_of_peft.tex
\begin{figure*}[htbp]
\centering
\subfigure[]{
    \hspace*{-1cm}
    \includegraphics[width=8cm]{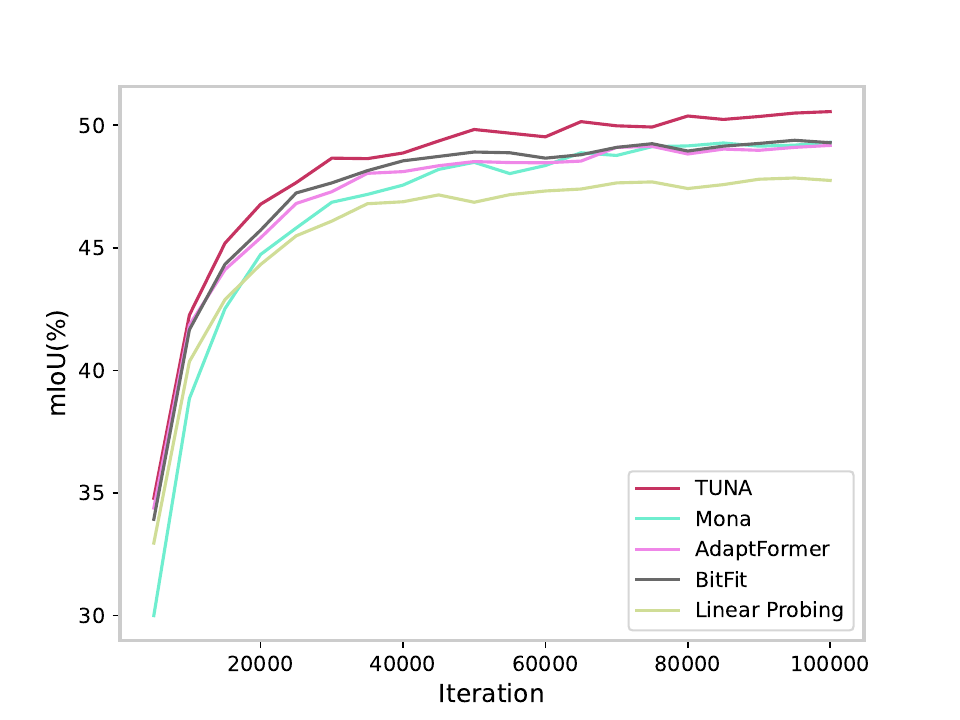}
    \label{fig:foodseg103}
}
\hspace*{-1cm}
\subfigure[]{
    \includegraphics[width=8cm]{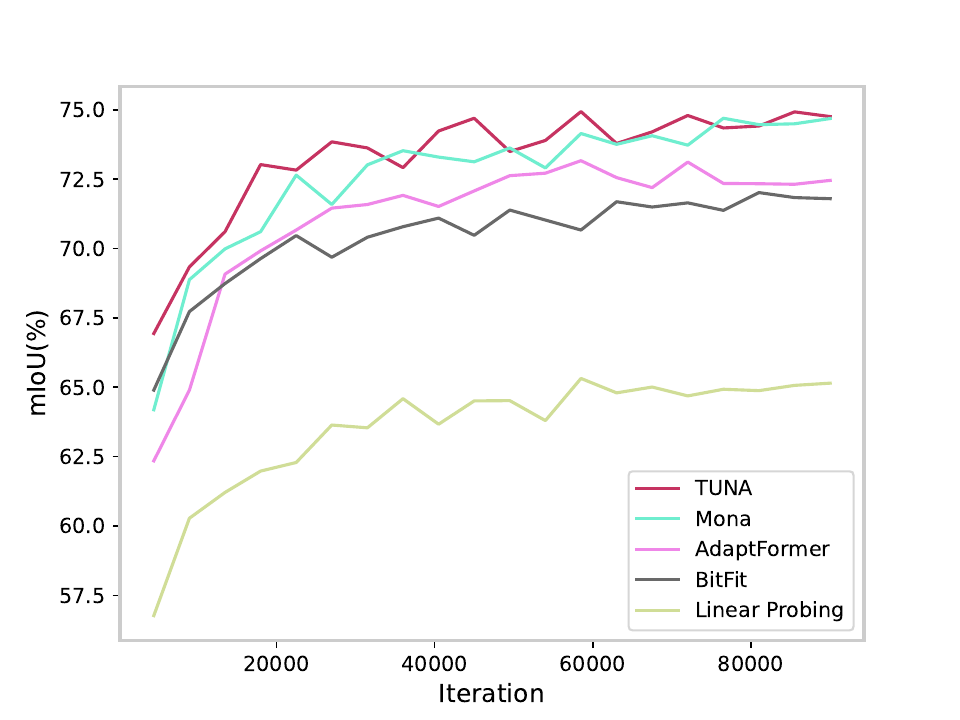}
    \label{fig:uec}
    \hspace*{-1cm}
}

\caption{Comparison of \mynet and other PEFT models on FoodSeg103 and UECFoodPix Complete. The horizontal axis represents the number of training iterations, and the vertical axis represents mIoU.}
\end{figure*}

%% file: tables/comp_foodseg.tex
\begin{table}[ht]
    \centering
    
    \scalebox{0.9}{
    \begin{tabular}{c|c|c|c|c|c|c} 
    \hline
    \multirow{2}{*}{Methods} & \multicolumn{3}{c|}{Metrics}  & \multirow{2}{*}{Crop Size} & \multirow{2}{*}{\makecell{Batch \\ Size}} & \multirow{2}{*}{Params(M)}  \\ 

    \cline{2-4} & mIoU(\%) & mAcc(\%) & aAcc(\%)  & & & \\ 
    \hline
    \multicolumn{7}{c}{\textbf{FULL FINETUNING}} \\ 
        \hline
FoodSAM\cite{foodsam} & 46.42 & 58.27 & 84.10 & 768$\times$768 & - & 632.75 \\
    \hline
    FDSNet(Swin)\cite{ly} & 47.34 & 60.04 & - & 768$\times$768 & 8 & 101.93 \\
    \hline
Swin-B & 41.20 & 53.90 & - & 512$\times$1024 & - & 88.52 \\
    \hline
deeplabV3+ & 36.22 & 48.87 & - & 512$\times$512 & - & - \\
    \hline
SeTR & 45.10 & 57.54 & 83.53 & 768$\times$768 & - & - \\

    \hline
    \multicolumn{7}{c}{\textbf{PEFT}} \\ 
        \hline
\textbf{Swin-TUNA} & 50.56 & 63.16 & 85.32 & 640 $\times$ 640 & 4 & 8.13 \\
    \hline
Mona & 49.31 & 62.07 & 84.79 & 640 $\times$ 640 & 4 & 5.27 \\
    \hline
BitFit & 49.39 & 61.97 & 84.89 & 640 $\times$ 640 & 4 & 0.50 \\
    \hline
AdaptFormer & 49.18 & 61.89 & 84.76 & 640 $\times$ 640 & 4 & 2.32 \\
    \hline
Linear Probing & 47.85 & 59.69 & 84.00 & 640 $\times$ 640 & 4 & 0 \\

    \hline
    \end{tabular}
    }
    \caption{Comparison with SOTA methods on FoodSeg103. The short horizontal line represents that it is not open source or not mentioned in the original paper}
    \label{tb.foodseg103}
\end{table}

%% file: tables/comp_uec.tex
\begin{table}[ht]
    \centering
    
    \scalebox{0.9}{
    \begin{tabular}{c|c|c|c|c|c|c} 
    \hline
    \multirow{2}{*}{Methods} & \multicolumn{3}{c|}{Metrics}  & \multirow{2}{*}{Crop Size} & \multirow{2}{*}{\makecell{Batch \\ Size}} & \multirow{2}{*}{Params(M)}  \\ 

    \cline{2-4} & mIoU(\%) & mAcc(\%) & aAcc(\%)  & & & \\ 
    \hline
    \multicolumn{7}{c}{\textbf{FULL FINETUNING}} \\ 
        \hline
FoodSAM\cite{foodsam} & 66.14 & 78.01 & 88.47 & 768$\times$768 & - & 632.75 \\    
    \hline
    FDSNet(Swin)\cite{ly} & 75.89 & 86.29 & - & 768$\times$768 & 8 & 101.93 \\
    \hline
GourmetNet\cite{GourmetNet} & 62.88 & 75.87 & 87.07 & 512$\times$512 & - & - \\
    \hline
\makecell{Bayesian \\ Deeplabv3+\cite{Bayesian}} & 64.21 & 76.15 & 87.29 & 320$\times$320 & - & - \\
    \hline
deeplabV3+ & 65.61 & 77.56 & 88.20 & 512$\times$512 & - & 54.74 \\
    \hline
CANet\cite{canet} & 68.90 & 80.60 & - & 640 $\times$ 640 & 4 & - \\
    \hline
\makecell{PSPnet(Fine-tuned \\ on Food2K)\cite{food2k}} & 74.50 & 84.10 & - & 640 $\times$ 640 & 4 & - \\

    \hline
    \multicolumn{7}{c}{\textbf{PEFT}} \\ 
        \hline
\textbf{Swin-TUNA} & 74.94 & 84.58 & 91.08 & 512 $\times$ 512 & 8 & 8.13 \\
    \hline
Mona & 74.70 & 83.89 & 91.04 & 512 $\times$ 512 & 8 & 5.27 \\
    \hline
AdaptFormer & 73.17 & 82.78 & 90.33 & 512 $\times$ 512 & 8 & 2.32 \\
    \hline
BitFit & 72.02 & 81.87 & 90.09 & 512 $\times$ 512 & 8 & 0.50 \\
    \hline
Linear Probing & 65.32 & 76.80 & 88.08 & 512 $\times$ 512 & 8 & 0 \\

    \hline
    \end{tabular}
    }
    \caption{Comparison with SOTA methods on UECFoodPix Complete. The short horizontal line represents that it is not open source or not mentioned in the original paper}
    \label{tb.uec}
\end{table}

%% file: figures/module_position.tex
\begin{figure}[ht]
    \centering
    \subfigure[]{
        \label{fig:structure}
        \includegraphics[width=7cm]{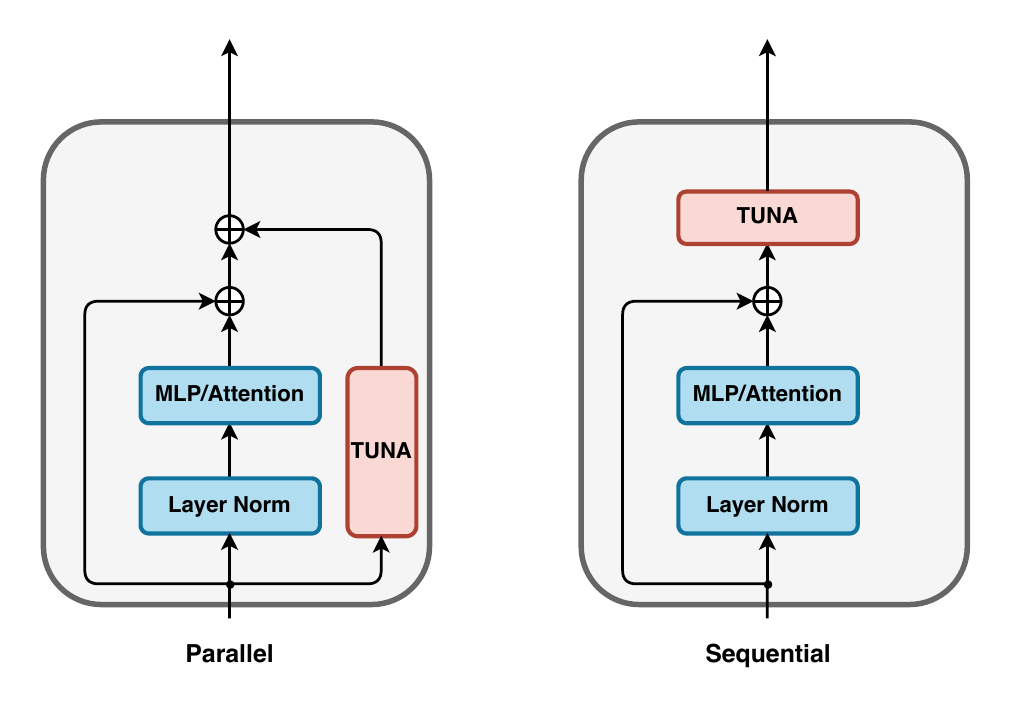}
    }
    \subfigure[]{
    
    \includegraphics[width=7cm]{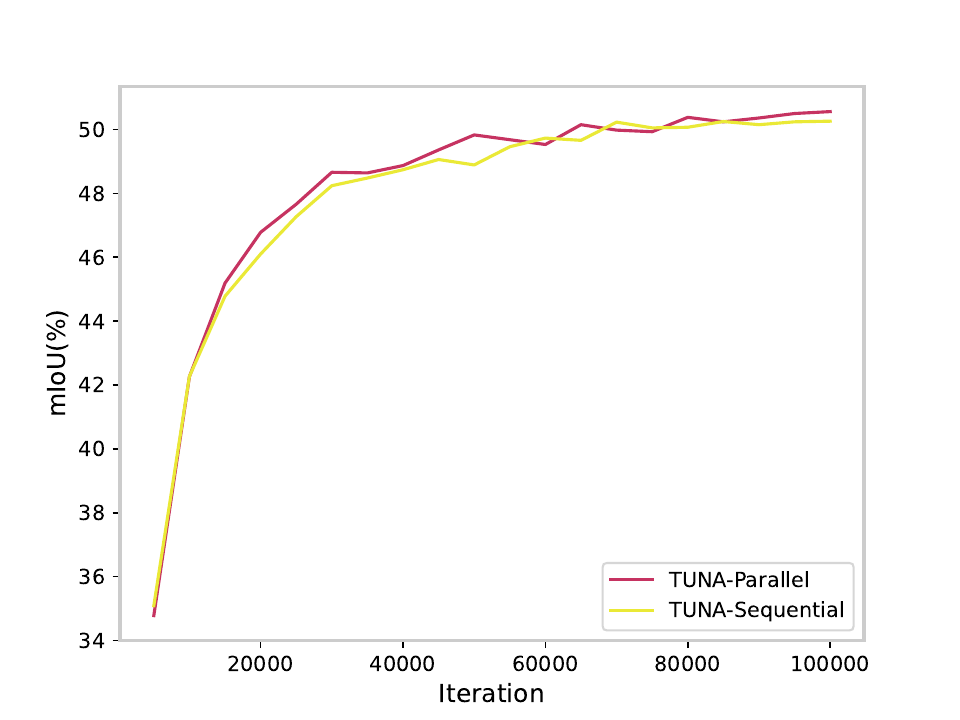}
        \label{fig:structure_res}
    }
    \caption{\ref{fig:structure} Two structures of the proposed method. The left side is the parallel structure and the right side is the parallel structure. \ref{fig:structure_res} The mIoU curves for the two different architectures during training. The two architectures demonstrate comparable performance, with the parallel structure exhibiting a slight advantage over the linear structure, achieving an accuracy of 0.3\% more than the latter.}
\end{figure}

%% file: tables/ablation_foodseg.tex
\begin{table}
    \centering
    \scalebox{0.9}{
    \begin{tabular}{c|c|c|c|c} 
        \hline
        \multicolumn{2}{c|}{Layer Related Parameters}                                                                                                                            & \multicolumn{3}{c}{Metrics}                                                                   \\ 
        \hline
        \multicolumn{1}{l|}{\begin{tabular}[c]{@{}l@{}}Adaptive~\\Convolution\end{tabular}} & \multicolumn{1}{l|}{\begin{tabular}[c]{@{}l@{}}Adaptive~\\Embedding~\end{tabular}} & \multicolumn{1}{l|}{mIoU(\%)} & \multicolumn{1}{l|}{mAcc(\%)} & \multicolumn{1}{l}{aAcc(\%)}  \\ 
        \hline
        \XSolidBrush                                                                        & \XSolidBrush                                                                       & 49.94                         & 62.55                         & 85.14                         \\ 
        \hline
        \Checkmark                                                                          & \XSolidBrush                                                                       & 49.88                         & 62.7                          & 85.18                         \\ 
        \hline
        \XSolidBrush                                                                        & \Checkmark                                                                         & 49.95                         & 62.57                         & 85.13                         \\ 
        \hline
        \Checkmark                                                                          & \Checkmark                                                                         & 50.56                         & 63.16                         & 85.32                         \\
        \hline
    \end{tabular}
    }
    \caption{Results of \mynet ablation experiments on different layer related parameters on FoodSeg103}
    \label{tb.alb}
\end{table}

%% file: sections/6_conclusion.tex
\section{Conclusion}
In this paper, we propose Swin-TUNA, a lightweight food semantic segmentation module based on PEFT. The primary objective of \mynet is to address the challenges posed by the substantial number of model parameters and the significant computational demands associated with food image segmentation tasks. The integration of a multi-scale adapter module with a hierarchical feature adaptation mechanism enables \mynet to achieve fine-grained segmentation of complex food images. This process is accomplished by updating a mere 4 percent of the total number of parameters while preserving the parameters of the pre-trained backbone network. Experiments demonstrate that the proposed method exhibits a substantial enhancement in performance when compared to existing full-parameter fine-tuning models. On the FoodSeg103 and UECFoodPix Complete datasets, the method achieves mIoU percentages of 50.56\% and 74.94\%, respectively.

Subsequent analysis indicates that Swin-TUNA's Hierarchical Feature Adaptation technique effectively captures high-frequency texture details with global semantic information in food images. Furthermore, the method exhibits augmented learning capability in small datasets scenarios, thereby providing a viable lightweight solution for real-time quality monitoring, composition analysis, and other applications in the food industry. 